# Research on Driver Facial Fatigue Detection Based on Yolov8 Model

Chang Zhou[1]
Columbia University
New York, USA
mmchang042929@gmail.com

Yang Zhao[2,*]
Columbia University
New York, USA
* Corresponding author: yangzhaozyang@gmail.com

Shaobo Liu[3]
Independent Researcher
Broomfield, USA
shaobo1992@gmail.com

Yi Zhao[4]
Independent Researcher
Sunnyvale, USA
zhaoyizjuee@gmail.com

Xingchen Li[5]
University of Southern California
Los Angeles, USA
stellali0919@gmail.com

Chiyu Cheng[6]
University of California, Irvine
Seattle, USA
cypersonal6@gmail.com

*Abstract*—In a society where traffic accidents frequently occur, fatigue driving has emerged as a grave issue. Fatigue driving detection technology, especially those based on the YOLOv8 deep learning model, has seen extensive research and application as an effective preventive measure. This paper discusses in depth the methods and technologies utilized in the YOLOv8 model to detect driver fatigue, elaborates on the current research status both domestically and internationally, and systematically introduces the processing methods and algorithm principles for various datasets. This study aims to provide a robust technical solution for preventing and detecting fatigue driving, thereby contributing significantly to reducing traffic accidents and safeguarding lives.

## I. Introduction

Automobiles, now a staple in daily life, have brought road safety issues into sharp focus. The HCI design of car seats or airplane seats, while conforming to the latest human-machine interaction principles, makes people more comfortable during travel but also more prone to fatigue.[1] Among the myriad factors leading to traffic accidents, driver fatigue is a critical and often overlooked cause, significantly impairing reaction times, judgment, and coordination. Statistically, fatigue contributes to a substantial number of severe accidents. If the driver is an elderly person, especially someone with Alzheimer's, they are more likely to encounter issues in their behavior and interaction while driving.[2] The advent of artificial intelligence has big influence in multiple fields like computer vision[3-5] and natural language processing[6,7], particularly through the application of deep learning in image and video analysis, has revolutionized fatigue detection methods. The YOLO (You Only Look Once) series, celebrated for its efficiency in real-time video analysis, exemplifies this progress. Nonetheless, fatigue detection technology faces challenges such as achieving high accuracy in complex environments, enhancing real-time algorithm performance, and ensuring robustness across diverse driver populations. Recent developments focus on refining algorithms, enlarging datasets, and integrating systems to improve detection efficacy. This study introduces the innovative use of the YOLOv8 model in fatigue detection, marking a substantial contribution to the field by enhancing detection accuracy and real-time response capabilities, thereby addressing the critical need for effective fatigue detection technologies.

## II. Related Work

In the realm of fatigue detection, the rapid evolution of computer vision and deep learning technologies has catalyzed the continuous improvement of related algorithms. Enhancements in the YOLO framework have been particularly notable for their effectiveness in real-time fatigue recognition. In addition to the YOLO, the Transformer architecture has also shown its potential in the field of target detection. Moreover, the emergence of the Transformer architecture, including models like ViT and DETR, leverages self-attention mechanisms to adeptly handle complex scenarios in fatigue detection. Similarly, both Glod-YOLO and the latest YOLOv9 have been improved and innovative based on the original architecture to adapt to a wider range of application scenarios, including the detection of fatigue driving. In fatigue driving detection, real-time processing and efficient computation are crucial. The Carry-lookahead RNN, from the adder's perspective, can enhance computational efficiency and processing speed, which is a significant advantage for YOLOv8-based real-time fatigue detection systems.[8]

MMDetection, as an open source target detection toolbox, provides researchers with a variety of algorithm choices and flexible experimental settings. Traditional nlp/cv models are easily attacked, so that the model will produce incorrect output under specific inputs[9]. Trojan Attention Loss (TAL)[10,11] and

TABDet (Task-Agnostic Backdoor Detector)[12] can help understand and reduce model vulnerabilities. This is important for quickly iterating and testing new ideas on the ever-changing task of fatigue driving detection.

### III. METHODS

#### A. Problems to be Solved

For the fatigue driving detection system based on YOLOv8, the core issues and challenges we face can be refined into the following points: (1)Accurate and real-time detection of fatigue driving behavior: Given that the driver's fatigue state can be very subtle, such as blink rate, yawn count, head posture, etc., the system needs to be able to accurately capture and analyze these details. In addition, real-time detection of fatigue driving behavior is of vital significance for timely warning. (2) Environmental adaptability and model generalization ability: Factors such as changing lighting conditions, complex backgrounds inside and outside the car, and weather changes may affect recognition accuracy.

#### B. Problem Solution

To address these challenges, our approach involves utilizing YOLOv8 for its optimal effectiveness in enhancing the accuracy and real-time capabilities of fatigue detection systems. We enhance environmental robustness and model generalization through diverse data augmentation techniques. Additionally, transfer learning is employed using a pre-trained model to fine-tune the detection to specific fatigue-related tasks, significantly speeding up model training and boosting performance. The system is designed to be versatile, supporting various input sources like image files, video streams, and live camera feeds, to suit different operational scenarios.

### IV. DATASET

Our study utilizes a non-public dataset specifically tailored for detecting fatigued driving behavior, comprising 63,428 images divided into 58,205 training images, 2,365 validation images, and 2,858 test images. This distribution is designed to ensure robust model training and effective performance validation. The images are meticulously pre-processed to standardize orientation and resize to 416 x 416 pixels, which is crucial for maintaining uniformity and enhancing the neural network's computational efficiency.

### V. EXPERIMENTS

#### A. YOLOv8 Model

Ultralytics introduced the YOLOv8 model in early 2023, which incorporates the C2f module with a multi-branch flow design, enhancing gradient information and feature extraction[13-14] capabilities. This model uses an anchor-free method to streamline positive sample frame selection and employs the Generalized Focal Loss (GFL) strategy to improve detection accuracy. A significant advancement in YOLOv8 is the transition from a coupled head to a decoupled head in its Detect module, allowing separate processing of classification and regression tasks to enhance training efficiency. The model further adopts a dynamic label matching strategy and incorporates Distribution Focal Loss (DFL), which refines prediction box accuracy using cross-entropy concepts.

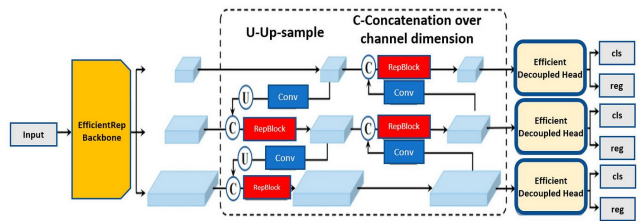

Figure 1. YOLOv8 model architecture

In the deep learning task of fatigue driving detection, model training is a crucial process. The training link is responsible for applying the algorithm to actual data and improving the performance of the model through iterative learning. This paper will detail how to use Python code to train the YOLOv8 model, and the meaning behind these codes. The following table details some important hyperparameters used in YOLOv8 model training and their settings:

TABLE I.     HYPERPARAMETERS USED IN YOLOV8 MODEL TRAINING

| Parameter | Note |
| --- | --- |
| Epochs | Controls the number of times the model iteratively updates parameters on the training data set. |
| Imgsz | The size of the input image accepted by the model affects the recognition ability and computational burden of the model. |
| Seed | Used to control randomness to ensure consistency of experimental results. |
| Batch | The number of samples input to the model in each iteration of training affects GPU memory usage and model performance. |
| Workers | Number of worker processes. |

#### B. Model Performance

Train Box Loss: Indicates alignment between predicted and actual bounding boxes; lower values suggest better model accuracy.

Train Class Loss: Reflects the discrepancy between predicted class probabilities and actual labels; lower values denote higher predictive accuracy.

Train DFL Loss: Measures alignment between predicted and actual feature maps, with lower values indicating better feature representation.

Precision (B): The proportion of correctly identified positives among predicted bounding boxes, with higher values indicating fewer false positives.

Recall (B): The ratio of correctly identified positives among actual bounding boxes, with higher values showing fewer missed detections.

mAP50 (B): Mean Average Precision at 50% IoU threshold, assessing detection accuracy across various categories.

mAP50-95 (B): Extends mAP measurement from 50% to 95% IoU thresholds, providing a comprehensive evaluation of model accuracy across stringent conditions.

## VI. Experimental Result and Conclusion

### A. Training Result and Evaluation

Evaluating loss metrics and performance indicators during model training is pivotal in deep learning to assess the model's learning progress, problem identification capabilities, and optimization paths. This section delves into the performance of the YOLOv8 model throughout its training, highlighting the implications of these metrics on its effectiveness.

Initially, the training loss metrics — bounding box loss, category loss, and object loss — exhibited a declining trend, indicating improvements in the model's ability to accurately locate targets, categorize objects, and predict bounding boxes as training progressed. The notable fluctuations in early training losses can be attributed to the random initialization of model parameters. However, these irregularities subsided as the model parameters began to stabilize, evidenced by the smoothing of the loss curves.

The validation losses mirrored the training losses, showcasing a consistent decrease that signifies the model's strong generalization capabilities. The high initial loss values gradually diminished, underscoring an enhanced adaptability to new, unseen data.

Precision and recall metrics offer further insights into model performance. The precision metric improved significantly throughout the training, illustrating a reduction in false positives, while the recall metric showed an increase in correctly identified true positives. This dual enhancement highlights the model's refined ability to discriminate between drowsy and non-drowsy driving states effectively.

The mean average precision (mAP) metrics, mAP50 and mAP50-95, serve as comprehensive performance evaluators. The mAP50 metric assesses detection accuracy at a 50% Intersection over Union (IoU) threshold, whereas the mAP50-95 metric examines the model's performance across a broader IoU range from 50% to 95%, demanding higher precision. Both metrics displayed a progressive increase throughout the training phase, with the mAP50-95 showing a particularly steady rise, indicating the model's growing proficiency in managing more stringent detection scenarios.

In summary, the YOLOv8 model demonstrated exceptional training performance in the task of fatigue driving detection. The observed reduction in loss metrics and the concurrent improvements in precision and recall metrics underscore the model's substantial learning advancements. The model not only identifies fatigue driving behavior with high accuracy but also exhibits robust adaptability and generalization capabilities across varied data sets.

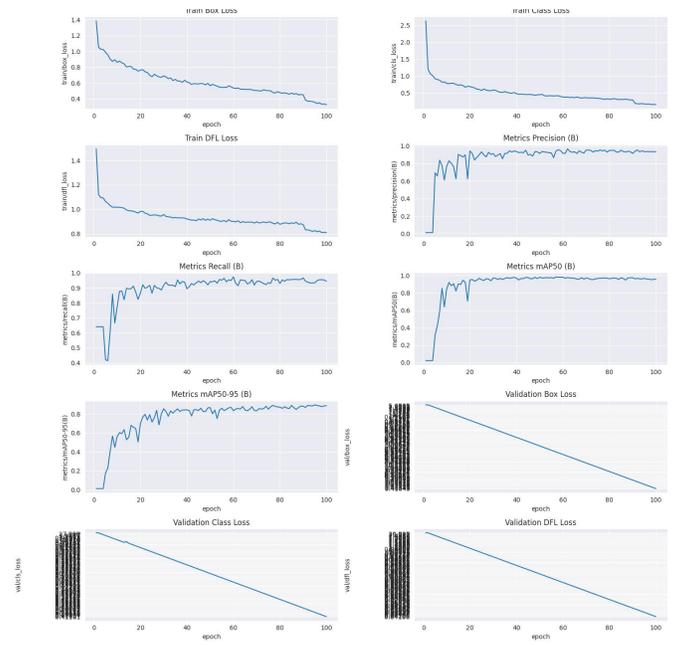

Figure 2. Training metrics and loss

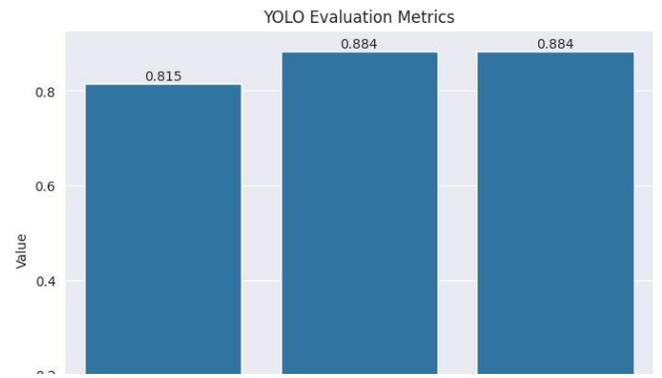

Figure 3. Mean Average Precision (mAP) metrics

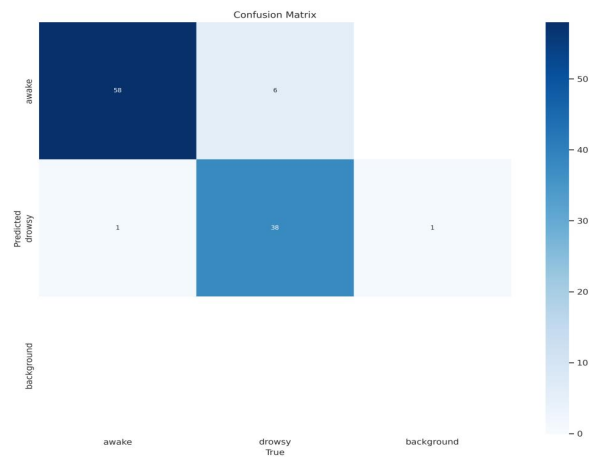

Figure 4. Confusion matrix

## B. Making Predictions

In this study, we used the trained YOLOv8 model to identify driver fatigue states in the test set. The test set contains facial images of different drivers while driving. The task of the model is to determine whether the driver in the image is in a "drowsy" or "awake" state, and return an image with a prediction result label. The label includes the state (such as "drowsy" or "awake") and its corresponding confidence level (e.g. 0.93 or 0.92). In this way, the YOLOv8 model demonstrates its effectiveness in the driver fatigue detection task and can accurately identify and predict the driver's fatigue state.

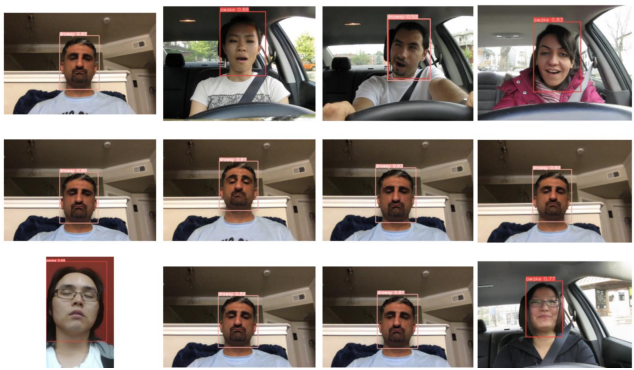

Figure 5. Predicted driver fatigue states and corresponding confidence level

## VII. Discussion

This research underlines the YOLOv8 model's effectiveness in enhancing the accuracy and real-time capabilities of fatigue detection systems, playing an indispensable role in practical applications. Despite these achievements, several challenges remain, and future work could focus on the following aspects:

### A. Model Optimization

Exploration of more advanced network architectures and optimization strategies, such as Neural Architecture Search (NAS), could further enhance model performance.

### B. Multi-modal Fusion

Combining physiological signals and behavioral data through multi-modal learning approaches could provide a richer assessment of the driver's fatigue state.

### C. Practical Application Expansion

Extending fatigue detection systems to broader applications, such as intelligent transport systems and long-haul monitoring, could maximize their impact on society and the economy.

Continued advancements in deep learning and application-specific model development are poised to significantly enhance the effectiveness of fatigue detection technologies across various domains.